\pdfoutput=1

\documentclass[11pt]{article}

\usepackage[]{ACL2023}
\usepackage{booktabs}
\usepackage{times}
\usepackage{latexsym}
\usepackage{graphicx} 
\usepackage{subcaption}
\usepackage[T1]{fontenc}

\usepackage[utf8]{inputenc}

\usepackage{microtype}
\usepackage{tabularx} 
\usepackage{inconsolata}
\usepackage{verbatim}
\usepackage{hyperref}
\usepackage{multirow}
\usepackage{svg}
\usepackage{listings}

\title{DACIP-RC: Domain Adaptive Continual Instruction Pre-Training via Reading Comprehension on Business Conversations\\
\texorpdfstring{%
  \raisebox{-0.2\height}{\includegraphics[height=7mm]{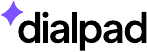}}%
}{}
}

\author{Elena Khasanova$^*$, Harsh Saini$^*$, Md Tahmid Rahman Laskar$^*$, Xue-Yong Fu$^*$, \\ \textbf{Cheng Chen}, \textbf{Shashi Bhushan TN} \\
\large{Dialpad Inc.} \\
  \texttt{\{xue-yong,elena.khasanova,tahmid.rahman,hsaini,cchen,sbhushan\}@dialpad.com}}

\begin{document}
\maketitle

\def\thefootnote{*}\footnotetext{\textbf{Equal Contributions. Sorted by the First Name.}}\def\thefootnote{\arabic{footnote}}

\begin{abstract}

The rapid advancements in Large Language Models (LLMs) have enabled their adoption in real-world industrial scenarios for various natural language processing tasks. However, the high inference cost of large-scale LLMs makes their deployment impractical, necessitating the use of smaller models. Despite their efficiency, smaller LLMs lack robust zero-shot instruction-following capabilities across diverse domains, limiting their adaptability to dynamic user requirements. Traditional fine-tuning approaches exacerbate this issue by inducing catastrophic forgetting, reducing the model’s generalization ability for unseen tasks. In this paper, we propose Domain Adaptive Continual Instruction Pre-Training via Reading Comprehension (DACIP-RC), a continual pre-training technique that enhances smaller LLMs' domain adaptability for business conversational tasks. Unlike conventional pre-training approaches that rely on next-token prediction, DACIP-RC generates diverse task instructions and responses via reading comprehension on conversation transcripts, enabling better instruction generalization. Our empirical evaluations demonstrate that DACIP-RC significantly improves zero-shot generalization across a wide range of business conversational tasks, including meeting summarization, action item generation, and call purpose identification. To the best of our knowledge, this is the first work to apply instruction pre-training on business conversational data, providing insights into how industries can leverage proprietary datasets for domain adaptation.

\end{abstract}
\section{Introduction}

Recently, there has been a huge increase in the industrial adoption of Large Language Models (LLMs) to build real-world NLP features \cite{laskar-etal-2023-building,fu-etal-2024-tiny,urlana2024llms}.  This is particularly due to the in-context learning capability of LLMs, which allows them to follow user-specified instructions to solve diverse tasks without requiring any fine-tuning on task-specific data \cite{ouyang2022training,zhao2023survey}. This is a major improvement over early transformer-based language models \cite{bart,pegasus,t5,DBLP:conf/nips/VaswaniSPUJGKP17,devlin2019bert} that did not have such instruction-following capabilities \cite{zhang2023instruction}. More importantly, this capability has revolutionized NLP, with the release of numerous LLMs having instruction-following capabilities achieving impressive zero-shot performance across a wide range of tasks \cite{laskar2023systematicchatgpt,qin2023chatgpt,bang2023multitask}. 

Despite the rapid progress in LLM development, the practical use of LLMs in real-world situations is hindered by their significant inference costs \cite{fu-etal-2024-tiny}. These costs increase with the size of the model, even though larger models offer superior instruction-following and zero-shot generalization abilities \cite{wan2023efficient,wang2024comprehensive}. 
Since the smaller models do not possess the same level of zero-shot instruction-following capabilities as their larger counterparts, 
it is important to adapt the smaller LLMs to the specific domain in which they will be deployed. 

One straightforward way in this regard could be to fine-tune  \cite{zhang2023instruction} smaller LLMs in task-specific datasets. 
However, fine-tuning LLMs to solve only certain tasks would limit their capability if there is a change in user requirements in the real world. For instance, suppose a model is just fine-tuned to generate a concise summary of a given conversation, but in the real world, the user may require the LLM to generate a bullet point summary. Fine-tuned smaller LLMs are also more prone to catastrophic forgetting and so they often lose their zero-shot instruction following capability after fine-tuning \cite{luo2023empirical,huang2024mitigating}. Therefore, it is important to adapt smaller LLMs to the target domain in a way such that they can properly follow diverse user instructions 

In this paper, we study how to effectively 
adapt smaller LLMs ($\leq$ 8B parameters) on real-world business conversational tasks such that they can achieve zero-shot generalization to perform diverse tasks (e.g. meeting summarization and action item generation, identifying the purpose of calls, etc.) depending on diverse instructions (e.g., long vs short vs bullet point summary generation, responding to multi-query prompts \cite{laskar-etal-2024-query}, etc.). While continual learning via next token prediction (NTP) \cite{wu2024continual} objective has demonstrated improved performance in the general domain, 
applying the NTP objective on noisy conversational data 
may not be useful to teach LLMs about conversation related tasks since such data lacks task-related information (e.g., in a business meeting, participants may not ask each other to summarize the meeting.). Moreover,  \citet{cheng-etal-2024-instruction} also demonstrate that continual pre-training via NTP objective lacks generalization in general NLP tasks.

Inspired by the success of techniques like continual pre-training via reading comprehension-based instruction pre-training \cite{chengadapting} for domain adaptation in general NLP tasks, in this paper, we propose a continual pre-training technique using diverse instructions 
by applying reading comprehension on conversation transcripts. Our proposed technique, \textbf{D}omain \textbf{A}daptive \textbf{C}ontinual \textbf{I}nstruction \textbf{P}re-Training via \textbf{R}eading \textbf{C}omprehension (\textbf{DACIP-RC}) successfully adapts LLMs to diverse business conversational tasks, achieving impressive zero-shot generalizability. Our major contributions in this paper are summarized below:

\begin{itemize}
    \item While prior work mostly studied domain adaptation for NLP tasks on non-conversational data, this paper is the first to propose an effective continual pre-training approach 
via reading comprehension-based instruction pre-training for better domain adaptation in business conversational tasks. 
\item We describe the construction of our pre-training dataset in extensive detail to enable others to replicate our approach. Therefore, findings from our extensive experiments 
will help industries working with conversational data to get insights on how to effectively leverage their proprietary datasets for 
zero-shot generalizability across conversational tasks.
\end{itemize} 



\section{Related Work}
LLMs are initially trained on massive, publicly available internet datasets using the self-supervised NTP objective \cite{radfordgpt1,radford2019languagegpt2,gpt3}, where the model learns to predict the next word in a sequence. However, the data used for this initial training can be vastly different from the specialized, proprietary data used in real-world industry applications. A notable example of this discrepancy is highlighted by 
\citet{wu2023bloomberggpt}, who demonstrate that general-purpose LLMs often struggle to match their performance on public benchmarks when applied to specific real-world tasks since public benchmarks are more aligned with the data they were originally trained. 

To bridge this performance gap, a technique known as continual pre-training has emerged \cite{wu2024continual}. This involves taking an existing LLM and further pre-training it on large amounts of a company's own unlabeled internal data. This process, which also leverages self-supervised learning, helps to adapt the model to the specific nuances and vocabulary of a particular domain \cite{gururangan2020don}, as shown by recent studies \cite{labrak2024biomistral,wu2024pmc} 

Nevertheless, this self-supervised pre-training approach based on the NTP objective is not without its drawbacks. Recent research indicates that while NTP-based continual pre-training on a specific domain can enhance performance on tasks within that domain, it may lead to a decrease in the model's ability to generalize 
across diverse NLP tasks \cite{cheng-etal-2024-instruction,chengadapting}. To address this, \citet{cheng-etal-2024-instruction} demonstrated that instruction pre-training by generating synthetic task instructions from unlabeled datasets via applying techniques like reading comprehension \cite{chengadapting} could improve domain adaptation and generalization in general NLP tasks. However, prior research on instruction pre-training of LLMs is mostly limited to the general domain, with no prior research studying the effectiveness of domain adaptation via 
instruction pre-training on conversations. 


   \begin{figure*}[t!]
       \centering
       \includegraphics[width=\linewidth]{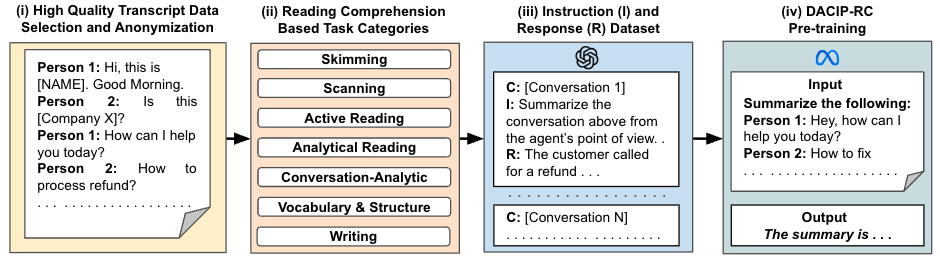}
       \caption{\small{An overview of our proposed DACIP-RC approach: (i) at first high high-quality transcripts are collected and anonymized, (ii) we demonstrate the reading comprehension tasks that were used to construct our (iii) instruction pre-training dataset  by generating instructions and responses using a closed-source LLM (e.g., GPT-4o-Mini \cite{openai2023gpt}) to (iv) pre-train a smaller open-source LLM (e.g., LLaMA-3.1-8B \cite{dubey2024llama3}).}}
     \label{fig:data_selection}
          
   \end{figure*}


 In this paper, we aim to address the gap in the prior research. Our focus is to investigate the effectiveness of continual instruction pre-training for domain adaptation across conversational tasks. We leverage large amounts of unlabeled business conversational data and apply reading comprehension-based techniques for question-and-answer generation from transcripts 
 for instruction pre-training.

 
 
 
 
\section{Methodology}
In this section, we present our DACIP-RC methodology to construct the instruction pre-training dataset using the reading comprehension technique. First, we demonstrate how we sample the data, followed by discussing how we construct instructions for pre-training inspired by the reading comprehension approach. Finally, we show how we generate and post-process the data to construct our instruction pre-training dataset. Figure \ref{fig:data_selection} summarizes our proposed 
methodology.
\subsection{Data Selection}
Our dataset consists of a large number of English-language transcripts from actual business conversations. These conversations, covering a range of topics, industries, and years, were transcribed using our proprietary, in-house automatic speech recognition (ASR) system. To ensure quality and diversity, we only used transcripts that were at least 120 seconds long, had high ASR confidence scores, and involved two or more speakers to reduce the likelihood of genres that are lacking diversity (e.g., voicemails or dropped calls).

We further utilize the approach discussed in \citet{xie2023efficient} and employ data selection based on token type entropy scores. We then post-process the transcripts by removing personally identifiable information and diversifying the formatting. We follow the approach discussed in \citet{zhang2024data} for data anonymization and use a combination of masking tokens (e.g. <COMPANY\_NAME\_1> instead of the real name) and noising tokens with custom, contextually relevant replacements (e.g. using a different person name) to allow the model to learn the properties of sensitive tokens without exposing them to the model. We diversify speaker tags (e.g. client vs sales representative) and introduce various transcript formats that include timestamps, different spacing configurations between the turns, etc. to ensure model robustness across different types of transcripts. 

\subsection{Pre-training Data Construction}
\subsubsection{Concept}
Inspired by the work on reading comprehension approaches to domain adaptation \cite{chengadapting, jiang2024improving}, as well as the theory and pedagogy behind reading comprehension, we designed a set of reading comprehension tasks aligned with various reading skills and activities commonly utilized in language and reading education \cite{brown2007teaching, Wright2001-WRICTA-4, CARRELL199747, BuchananHill19102016}. These tasks are structured around three primary objectives:

(i) Enhancing the model’s ability to discern the underlying structure of transcripts and accurately retrieve factual and relevant information in downstream tasks.

(ii) Increasing exposure to domain-specific knowledge, particularly within industry-specific business conversational contexts.

(iii) Bridging the gap between general instruction tuning and task-specific fine-tuning.

To accomplish these goals, we developed tasks across the following categories that correspond to four major types of reading:

\textbf{(i) Skimming}: Tasks related to big-picture understanding of the conversation. \\
Examples: \textit{What is the main topic of this conversation? What are the stated goals of the speakers in this dialogue?}

\textbf{(ii) Scanning}: Tasks aimed at extracting specific details from the text. \\
Examples: \textit{When did Alfildr indicate that the email confirmation would be sent? The meeting is scheduled at \_\_\_ PM next Wednesday.}

\textbf{(iii) Active Reading}: Tasks related to actively engaging with the text such as taking notes, summarizing, asking questions, or retrieving information. \\
Example: \textit{Identify topics discussed in this call and write a two-sentence summary of each topic. Then, rewrite this conversation as a marketing email.}

\textbf{(iv) Analytical Reading}: Tasks related to discussing underlying assumptions, potential biases, evidence, perspectives of the speakers in the conversation, and various types of divergent questions typically without the right answer.\\
Example: \textit{Why did the prospect reject the sales agent’s proposal? Discuss possible reasons and potential solutions. Explain your rationale based on the provided transcript.}

We also added three more categories that overlap with the reading techniques listed above but have a more specialized focus:

\textbf{(v) Conversation-Analytic tasks}: Tasks related to underlying conversational structure and how it is realized in the transcript, e.g. questions addressing turn-taking, organization of utterances, utterance intent, handling misunderstandings. \\
Example: \textit{Identify questions under discussion (QUD). Analyze the relevance of Speaker 1's inquiries. How did they resolve the QUD?}

\textbf{(vi) Vocabulary and Structure}: Tasks related to terminology, structure, and composition of the transcript. \\
Example: \textit{How is the term 'open balance' defined in financial terms? Reorder the following conversation sections to reflect the correct sequence: A. Computer Issues Discussion. B. Final Arrangements. C. Personal Updates and Plans. }

\textbf{(vii) Writing}: Tasks related to text generation tailored to specific industries and genres of business writing to enhance the representation of in-domain vocabulary that typically do not rely on the transcript in the instruction. \\
Example: \textit{Create a dialogue between a client and a project manager in the Translation Services industry that addresses these topics: Billing Preferences, Current Context, Implementation Timeline.}

This approach combines the benefits of in-domain knowledge and instruction-following capabilities. Instead of learning a narrowly defined task, the model acquires intermediate competencies 
for completing diverse downstream tasks with enhanced transcript understanding.
\subsubsection{Generation Method}
\textbf{Generation:}
For each category described in the previous section, we curated a set of 41 meta prompts that we used to prompt an LLM (GPT-4o-Mini \cite{openai2023gpt}) to generate a series of tasks with corresponding answers using a given transcript. The prompts generally follow this template: 
\texttt{\{transcript\}\{meta-instruction\}\{output requirements\}\{output example\}}. See an example in Appendix \ref{appendix:a}.

For each category, we generate tasks of different varieties, where applicable, which include but are not limited to different types of questions (e.g. yes-no, multiple choice, open-ended), gap-filling and reordering tasks, information extraction, summarization, etc., which may require generating new text, returning a snippet from a transcript verbatim, or a combination of both. We also vary the number of tasks generated per transcript and the output length and format requirements (e.g. a letter, a paragraph, several words, bulleted list). 

We developed the meta prompts via extensive prompt-engineering. To minimize API costs and improve generation quality, these meta prompts are designed to generate multiple questions/task instructions per transcript and their corresponding responses in JSON format to ensure easier parsing as seen in Appendix \ref{appendix:a}.

\textbf{Parsing:} Custom parsers are implemented for different types of output to reliably extract the task instructions and the answers separately. We remove responses that failed to parse due to invalid JSON outputs. After this step, we keep 10\% of the generated data unparsed and use original meta prompts with their outputs in JSON format 
to improve the model's JSON format following capability. 
\begin{table*}[t]
\setlength{\tabcolsep}{3pt}
\centering
\tiny
\begin{tabular}{lcccccccccc}
\toprule
\multirow{2}{*}{\textbf{Model}}& 
\multicolumn{3}{c}{\textbf{Text Classification (F1-Score)}} &
\multicolumn{5}{c}{\textbf{Text Generation (ROUGE-2)}} & \multirow{2}{*}{\textbf{Overall Avg.}} \\
\cmidrule(lr){2-4} \cmidrule(lr){5-9}
 & {\textbf{Call Outcome}} & {\textbf{PoC Category}} & \textbf{Avg.} & {\textbf{PoC Explanation}} & {\textbf{Action Items}}  & \textbf{Support Call Sum.} & \textbf{Meeting Sum.} &  \textbf{Avg.} &
 \\
\midrule
\textbf{LLaMA-3.1-8B-Instruct}         & 13.85 & 26.46 & 20.16 & 13.71 & 16.02 & 23.22 & 15.27  & 17.06 &  18.09 \\
\textbf{LLaMA-3.1-8B-DACIP-RC}       & 42.36  & 54.09 & 48.23 & 15.34  & 16.40 & 20.46 & 17.05 & 17.31 & 27.62 \\
\textbf{LLaMA-3.1-8B-Instruct-DACIP-RC}       & 46.23  & 53.31  & 49.77 & 14.08 & 23.89 & 19.46 & 17.68 & 18.78 & 29.11
\\
\bottomrule
\end{tabular}
\caption{\small{Performance on in-domain conversational tasks. Here, `Sum.' denotes `Summarization' and `Avg.' denotes `Average'.}}
\label{tab:internal_results}
\end{table*}

\textbf{Postprocessing:}
Out of the parsed task-response pairs, we construct prompts and outputs that we use for instruction pre-training. 70\% of the pairs are used for single-task prompts, and 30\% are used for multitask prompts, with up to 10 instructions for the same transcript. In multitask prompts, the tasks can fall into the same type (e.g. all are multiple choice "scanning" questions) or different categories, and may require outputs in the same or different formats. We also vary whether the instructions appear before or after the transcript. See Appendix \ref{appendix:b} for an example of a pretraining task.

The resulting dataset consists of over 26M instances, with an average prompt length (including transcript) of 
1448.46 tokens and a response length of 107.09 tokens. This adds up to roughly 25B tokens estimated using \textit{tiktoken}\footnote{\url{https://github.com/openai/tiktoken}} tokenizer. 
\section{Experiments}
\subsection{Training Settings}
The pre-training job was conducted using LLaMA-3.1-8B\cite{dubey2024llama3} (both base and instruct models) on 8 nodes, each with 8 x NVIDIA H100 80 GB GPUs attached.
Deepspeed \cite{aminabadi2022deepspeed} was used in conjunction with PyTorch to orchestrate the training, with relevant training parameters are as follows: datatype: bf16,  learning rate: $2e-6$, number of epochs: $1$.

\subsection{Evaluation Benchmarks}
\paragraph{Internal Benchmarks:}
Our internal benchmarks consist of the following four tasks (containing 1.5k instances in total): 

\textbf{(i) Action Items}: The task involves the generation of a list of actionable tasks extracted from a transcript of a conversation. Each task is a short description of an activity that should occur after the conversation has ended. 

\indent \textbf{(ii) Purpose of Call}: This task considers two cases: (i) classifying the main purpose of a conversation 
 to one of several pre-defined categories \cite{khasanova-etal-2022-developing} and (ii) free-form generation of the call purpose alongside an explanation. 

 \indent \textbf{(iii) Call Outcome}: The task aims to classify the main outcome of a conversation to one of several pre-defined categories.

 \textbf{(iv) Summarization}: The task is to generate a concise conversation summary according to specific requirements such as summary length (long, medium, or short) or format (e.g. bullet points). We consider two real-world cases for summarization: (i) summarizing customer-agent support calls, and (ii) summarizing business meetings. 

See Appendix \ref{appendix:c} for the prompt for each task.
\paragraph{External Benchmarks:}
For external benchmarking, we use the QMSUM dataset \cite{zhong2021qmsum}, which focuses on query-based meeting summarization and includes 281 test instances.
\subsection{Experimental Results}

\subsubsection{Performance on Internal Benchmarks}
We evaluate our proposed \textbf{DACIP-RC} based \textbf{LLaMA-3.1-8B} models against the baseline \textbf{LLaMA-3.1-8B-Instruct} model on both text classification and text generation tasks in zero-shot.
Table~\ref{tab:internal_results} summarizes the results across different tasks.

We observe that applying \textbf{DACIP-RC} leads to significant performance improvements on all classification tasks, with the average F-1 score more than doubling for both \textbf{LLaMA-3.1-8B-DACIP-RC} and \textbf{LLaMA-3.1-8B-Instruct-DACIP-RC} in comparison to the baseline. On text generation tasks, the \textbf{DACIP-RC} models also outperform the baseline on ROUGE-2 \cite{lin2004rouge} metric, with the exception of \textit{Support Call Summarization} with a slight performance drop. Particularly noteworthy is the performance gain on \textit{Action Items}, where \textbf{LLaMA-3.1-8B-Instruct-DACIP-RC} achieves a ROUGE-2 increase from 16.02 (baseline) to 23.89.

Based on the autometrics and upon closer inspection of the outputs, we conclude the following:
\indent (i) The advantages of \textbf{DACIP-RC} method are particularly evident on the tasks that rely on a detailed understanding of a transcript, such as \textit{Purpose of Call (PoC)}, \textit{Action Items}, \textit{Call Outcome};\\
\indent (ii) On summarization tasks, \textbf{DACIP-RC} brings more significant improvements on \textit{Meeting Summarization}, with longer and more complex multiparty conversations, than on shorter two-party dialogues in the \textit{Support Call Summarization}; \\
\indent (iii) \textbf{DACIP-RC} models favor shorter and more concise summaries than the baseline model, which can explain the drop in performance observed on \textit{Support Call Summarization}.


Overall, the \textbf{DACIP-RC} method consistently demonstrates its effectiveness in improving the performance across diverse in-domain conversational tasks, with the \textbf{LLaMA-3.1-8B-Instruct-DACIP-RC} model achieving the best overall performance (having an average score of 29.11).

\begin{table}[t]
\setlength{\tabcolsep}{4pt}
\centering
\scriptsize
\begin{tabular}{lcccc}
\toprule
{\textbf{Model}} & {\textbf{B-S}} & \textbf{R-1} & \textbf{R-2} & \textbf{R-L} \\
\midrule
\textbf{LLaMA-3.1-8B-Instruct} & 36.72 & 11.16 & 3.12 & 7.73 \\
\textbf{LLaMA-3.1-8B-DACIP-RC} & 49.00 & 18.09 & 4.28 & 11.93 \\
\textbf{LLaMA-3.1-8B-Instruct-DACIP-RC} & 59.74 & 27.57 & 6.32 & 17.95 \\
\bottomrule
\end{tabular}
\caption{\small{Performance comparison on external benchmarks: the QMSUM \cite{zhong2021qmsum} dataset. Here, `B-S' denotes `BERTScore' \cite{zhang2019bertscore} and `R' denotes `ROUGE'.}}
\label{tab:external_results}
\end{table}


\subsubsection{Performance on External Benchmarks}
To evaluate the generalization ability of our models beyond in-domain tasks, we assessed their performance on the QMSUM dataset, a public benchmark for query-focused meeting summarization. 
Based on the results presented in Table~\ref{tab:external_results}, we find that the baseline \textbf{LLaMA-3.1-8B-Instruct} achieves the lowest scores across all metrics, with a ROUGE-1 of 11.16, ROUGE-2 of 3.12, and ROUGE-L of 7.73, along with a BERTScore of 36.72. 

This demonstrates that the \textbf{DACIP-RC} method helps both base and instruct \textbf{LLaMA-3.1-8B} models to achieve a huge gain in performance across all metrics, with the \textbf{LLaMA-3.1-8B-Instruct-DACIP-RC} model achieving the best result. 

     \subsubsection{Ablation Studies} 
\begin{figure}[t!]
    \centering
    \includegraphics[scale=0.4]{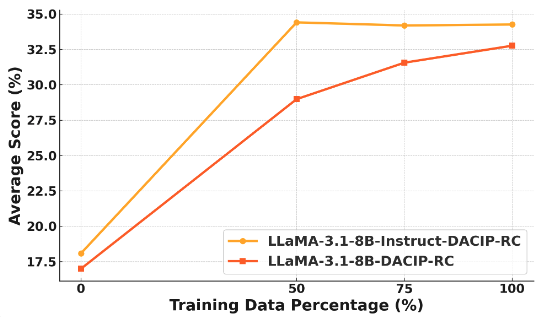}
    \caption{Ablation Tests on Training Data Size}
   
    \label{fig:scaling}
\end{figure}

Our ablation study on training data size (Figure \ref{fig:scaling}) shows that the base LLaMA-3.1 model's performance consistently improves with more data. The instruct model, however, is more data-efficient; it gets a significant initial boost, but performance gains become negligible as the dataset grows.

\subsubsection{Qualitative Evaluation}

We also conduct an LLM-judge-based reference-free evaluation using the \textit{Gemini-2.5-Pro} model as the judge on the text generation tasks across 500 test samples by considering factual correctness, adherence to instructions, format following, and conciseness. The results are as shown in Table \ref{tab:llm_judge_results}.

We find that in terms of \textit{pointwise} evaluation (Likert-scale, 1-5), the baseline \textbf{LLaMA-3.1-8B-Instruct} model got an average score of 2.08, while our \textbf{LLaMA-3.1-8B-Instruct-DACIP-RC} model achieved an average score of 4.07. Moreover, in terms of pairwise evaluation, the \textbf{LLaMA-3.1-8B-Instruct-DACIP-RC} model was preferred 85.2\% of the time. More interestingly, the DACIP-RC was preferred in 93.7\% of the cases in the support call summarization task, in which it performed slightly poorer than the baseline in reference-wise evaluation\footnote{Differences in performance between reference-free and reference-wise metrics in the evaluation of LLMs are quite common in the literature \cite{liu2023revisiting}.} in terms of ROUGE-2.  This evaluation further demonstrates the effectiveness of DACIP-RC.

\begin{table}[t!]
\centering
\tiny

\setlength{\tabcolsep}{2pt}
\renewcommand{\arraystretch}{1.05}
\begin{tabular}{lcc}
\toprule

\textbf{Model} & \textbf{Pointwise (Likert 1-5)} & \textbf{Pairwise Preference} \\ \hline
\textbf{LLaMA-3.1-8B-Instruct} & 2.08 & 13.7 \\
\textbf{LLaMA-3.1-8B-Instruct-DACIP-RC} & 4.07 & 85.2 \\

\bottomrule
\end{tabular}

\caption{\small{Reference-free evaluation using \textit{Gemini-2.5-Pro} across 500 test samples.}}
\label{tab:llm_judge_results}
\end{table}

\subsubsection{Out-of-Domain Generalization}

\begin{table}[t!]
\centering
\scriptsize
\begin{tabular}{@{}lcc@{}}
\toprule
 & \textbf{PubMedQA} & \textbf{MediQA-QS} \\
\midrule
\textbf{LLaMA-3.1-8B-Instruct }             & 55.0 & 7.76 \\
\textbf{LLaMA-3.1-8B-Instruct-DACIP-RC }    & 57.8 & 9.32 \\
\bottomrule
\end{tabular}
\caption{\small{Out-of-Domain Generalization Results on PubMedQA and MediQA-QS datasets, in terms of Accuracy and ROUGE-2, respectively.}}
\label{tab:ood}
\end{table}

While the focus of our work is to demonstrate how we adapt an LLM in a real-world industrial setting for business conversational tasks, we further evaluated our DACIP-RC model in the biomedical domain to investigate its generalizability and the presence of any catastrophic forgetting. We select one classification task, the PubMedQA \cite{jin2019pubmedqa} dataset for question answering, and one summarization task, the MediQA-QS \cite{abacha2021overview} dataset for healthcare question summarization. These two datasets are commonly used benchmarks in the biomedical domain for the evaluation of LLMs \cite{jahan2024comprehensive}. We compare the performance of DACIP-RC-tuned \textbf{LLaMA-3.1-8B-Instruct} with the original zero-shot \textbf{LLaMA-3.1-8B-Instruct} evaluated by \citet{jahan2025evaluating}. Based on the results given in Table \ref{tab:ood}, we find that our proposed DACIP-RC approach led to a performance gain even in the out-of-domain biomedical benchmarks on related tasks (e.g., question answering and summarization).  To further demonstrate that typical fine-tuning of LLMs only on a certain task could lead to catastrophic forgetting, we adopt the \textbf{Mistral-7B-Instruct} model \cite{jiang2023mistral} trained only on the QMSUM-Multi-Query \cite{zhong2021qmsum,laskar-etal-2024-query} dataset for Query-Focused Meeting Summarization from  \citet{laskar-etal-2024-query} and evaluate on our internal text classification datasets (Call Outcome and Purpose of Call Category Classification). We find that while the zero-shot \textbf{Mistral-7B-Instruct} achieves an average F1 score of 35.5, the Mistral model fine-tuned only on QMSUM performs very poorly in our classification datasets (average F1 score of 0.0). This demonstrates that typical task-specific fine-tuning can lead to catastrophic forgetting. 
\subsubsection{Performance Comparison with the Standard NTP Objective}
In our proposed DACIP-RC approach, we only predict the response for the given conversation. To investigate the effectiveness of this approach, we pre-train the \textbf{LLaMA-3.1-8B} model using the standard NTP objective on our DACIP-RC dataset, consisting of 25M conversations alongside the corresponding DACIP-RC instruction-response pairs. We denote this model as \textbf{LLaMA-3.1-8.B-Internal-NTP}. Unlike our DACIP-RC recipe (which only predicts the response), this baseline predicts the next token over the entire sequence, which includes the conversation, task instruction, and response. This is done because NTP pre-training only on conversations can destroy the model's zero-shot instruction-following abilities \cite{fu2025dacp}. We find that our \textbf{LLaMA-3.1-8B-DACIP-RC} model outperforms the \textbf{LLaMA-3.1-8.B-Internal-NTP} model by a large margin, with the average result across our internal datasets being 29.1 for DACIP-RC, while only 18.5 for the NTP baseline.

\subsubsection{Incorporating Structured Output Format for Real World Inference}
For real-world inference, generating outputs in a structured format is crucial \cite{laskar-etal-2024-query}. With structured output such as JSON, developers can parse the results reliably to retrieve the required information to power downstream tasks. Therefore, we further performed experiments to investigate whether our DACIP-RC model is compatible with structured output generation techniques. In this regard, we compare the plain decoding method and structured output generation by providing the inference engine (SGLang~\cite{zheng2024sglangefficientexecutionstructured}) a JSON schema to guide the generation of output tokens. We choose three tasks that require a JSON formatted output and provide their Pydantic~\footnote{\url{https://docs.pydantic.dev/latest/}}
based decoding hint as follows:

\textbf{(i) Combined summarization and action items}: This task prompts the model to generate a JSON object with two keys, ``summary" and ``action\_items". The JSON decoding hint is: \texttt{\textbf{RootModel[dict[str, Union[str, list[str]]]]}}

\textbf{(ii) Question Answering (QA)}: This task prompts the model to answer a list of questions about the transcript, each of which is represented by a key-value pair in the output JSON object. The JSON decoding hint is: \texttt{\textbf{RootModel[dict[str, list[str]]]}}


For this experiment, we only provide the JSON schema with minimal hints. From Table~\ref{tab:json_decoding_results}, we can see that even with a minimal hint of JSON-constrained decoding method, the model performance is greatly improved for all tasks.

\begin{table}[t]
\scriptsize
\setlength{\tabcolsep}{4pt}
\centering
\begin{tabular}{lccc}
\toprule
\textbf{Decoding Method}  & {\textbf{Summarization}} & {\textbf{Action items}} & {\textbf{QA}} \\ \midrule
\textbf{Plain decoding} & 0.1658 & 0.1672 & 0.46 \\ 
\textbf{JSON constrained decoding} &  0.1711 & 0.2674 & 0.7 \\ \bottomrule
\end{tabular}
\caption{\small{Performance comparison (ROUGE-2 for summarization and action items, F1-Score for QA) between causal language model decoding and JSON constrained decoding. }}
\label{tab:json_decoding_results}
\end{table}


\section{Conclusion}

This paper introduced DACIP-RC, a continual instruction pre-training method that enhances smaller LLMs' adaptability to business conversational tasks. By leveraging our proposed reading comprehension-based instruction generation from transcripts approach, DACIP-RC significantly improves zero-shot generalization in both text classification and generation tasks. 
Moreover, the proposed DACIP-RC framework relies on a one-time manual creation of 41 meta-prompts that automates the generation of over 25 million training instances.
Therefore, DACIP-RC offers a scalable approach to improve the performance of LLMs in real-world applications. 
In the future, we will explore the creation of new benchmarks to evaluate the DACIP-RC approach. 

\section*{Limitations}
The proposed reading comprehension-based instruction pre-training dataset construction approach is only evaluated on conversational data in the business communication domain. Thus, the findings from our experiments may not be applicable to other domains. 
Moreover, the analysis of the impact of various 
subsets of reading comprehension 
data on the downstream task performance is left out of the scope of this study. In the future, we plan to do ablation studies with various subsets of our datasets in search of a more optimal data mix for domain adaptation. Even though the proprietary dataset used in this paper cannot be released, our decision is also in full compliance with the EMNLP Industry Track policy. In addition, to ensure our research is transparent and replicable, we provided a detailed description of our DACIP-RC methodology throughout the paper so that other researchers can replicate our approach on their own conversational datasets. 


\section*{Ethics Statement}

We adhered to the licensing terms of the various tool providers we used (e.g., Meta, HuggingFace). Since this study involves the use of proprietary data containing sensitive information, we applied a rigorous anonymization process to safeguard sensitive details to ensure the privacy and security of the internal data. In line with established privacy best practices \cite{narayanan2007breakanonymitynetflixprize}, we have decided not to release these datasets publicly to fully mitigate any risk of sensitive information leakage. 

\bibliography{anthology,custom}
\bibliographystyle{acl_natbib}

\appendix

\section{Appendix}
\subsection{Meta Prompt Example}
\label{appendix:a}
A metaprompt example for a wh- question in the \textit{scanning} category. \\

{\fontfamily{qcr}\selectfont{\normalsize \small
=== Conversation Starts ===

\{conversation transcript\}

=== Conversation Ends ==== \\

Based on the conversation above, generate a series of 5 open-ended questions that start with Who, When, Where, What, Which, How, Why. Provide answers to these questions. Questions 1-3 should have an answer as a complete sentence or several sentences. Questions 4-5 should have a short answer of just a few words. Write the results in the JSON format with two keys: question, answer. 

Example output: 

[ { "question": "When are Person 1 and Person 2 planning to meet?" , "answer": "Wednesday at 2 PM" }, 
{ "question": "What time zone is Person 1 in?", "answer": "Person 1 is in the Pacific time. } ]

Be creative. Do not include the example output in your response. \\
}}
\subsection{Pre-training task example}
\label{appendix:b}

An example of a pretraining task after parsing and postprocessing:

{\fontfamily{qcr}\selectfont{\normalsize \small
=== Transcript Start ===

\{transcript\}

=== Transcript End ===

Provide responses for the following questions:\
1. Select all that apply to answer this question: What are some features discussed in the conversation? A. Installing a fan. B. Buying a TV. C. Building a recess for the fireplace. D. Adding a sound bar.

2. Which of the following is incorrect regarding the plan for the fireplace? A. There will be a TV above it. B. It will be part of a renovation. C. The fireplace is meant for outdoor use. D. The fireplace is meant to be recessed into the wall.

3. Select the correct option: What will be included in the follow-up email Person 1 is sending? A. Installation manual. B. Quote for the fireplace. C. Details about the fan. D. All of the above.
}}

\subsection{Prompt Format for Downstream Tasks}
\label{appendix:c}

\subsubsection{Summarization}
{\fontfamily{qcr}\selectfont{\normalsize \small Generate a summary of the \{Length Type\} following conversation \{Format\} without assessing its quality.
    }}\\
Note: length (long, medium, short) and format (e.g., bullet points) requirements in the prompt. 


\subsubsection{Action Items}
{\fontfamily{qcr}\selectfont{\normalsize \small For the conversation given below, generate a newline-separated list of work, business, or service-related TODO tasks that should be completed after the conversation. Each task is a one-sentence summary of the action to be taken.\\
Transcript: \{\}
    }}

\subsubsection{Call Outcome}
{\fontfamily{qcr}\selectfont{\normalsize \small For the conversation below, apply the appropriate category from the list provided below to describe the outcome of the conversation (respond with `Other' if no category applies): Call back,
Unsuccessful contact,
Voicemail Success,
Payment / Billing,
Status update,
Scheduled appointment,
Cancellation. \\
Transcript: \{\}
    }}\\
Note: this is not an exhaustive list of outcomes that are supported.

\subsubsection{Purpose of Call}
{\fontfamily{qcr}\selectfont{\normalsize \small For the conversation below, identify a single category for the purpose of the conversation chosen from this list: Account Management,
Appointment,
Billing Questions,
Callback,
Cancellation,
Claim,
Complaint.\\
Transcript: \{\}
    }}\\
Note: this is not an exhaustive list of supported categories.

\end{document}